\documentclass[pdflatex,sn-mathphys-num]{sn-jnl}% Math and Physical Sciences Numbered Reference Style 
\makeatletter
\renewcommand{\orcidlogo}{}%
\renewcommand{\orcid}[1]{}%
\providecommand{\orcidlink}[1]{}%
\makeatother
\usepackage{graphicx}%
\usepackage{multirow}%
\usepackage{amsmath,amssymb,amsfonts}%
\usepackage{amsthm}%
\usepackage{mathrsfs}%
\usepackage[title]{appendix}%
\usepackage{xcolor}%
\usepackage{textcomp}%
\usepackage{manyfoot}%
\usepackage{booktabs}%
\usepackage{algorithm}%
\usepackage{algorithmicx}%
\usepackage{algpseudocode}%
\usepackage{listings}%
\usepackage{multicol}
\usepackage{rotating, longtable, lscape}
\usepackage{fontawesome}

\begin{document}

\title[]{From Framework to Reliable Practice: End-User Perspectives on Social Robots in Public Spaces}


\author*[1,3]{\fnm{Samson} \sur{Oruma}}\email{samsonoo@hiof.no}

\author[2]{\fnm{Ricardo} \sur{Colomo-Palacios}}\email{ricardo.colomo@upm.es}

\author[3]{\fnm{Vasileios} \sur{Gkioulos}}\email{vasileios.gkioulos@ntnu.no}

\affil*[1]{\orgdiv{Department of Computer Science and Communication}, \orgname{Østfold University College}, \orgaddress{\street{Bra veien 4}, \city{Halden}, \postcode{1757}, \state{Viken}, \country{Norway}}}

\affil[2]{\orgdiv{Escuela Técnica Superior de Ingenieros Informáticos}, \orgname{Universidad Politécnica de Madrid}, \orgaddress{\street{C. de los Ciruelos}, \city{28660 Boadilla del Monte}, \postcode{28040}, \state{Madrid}, \country{Spain}}}

\affil[3]{\orgdiv{Department of Information Security and Communication Technology}, \orgname{Norwegian University of Science and Technology}, \orgaddress{\street{Teknologivegen 2}, \city{Gjøvik}, \postcode{2802}, \state{Innlandet}, \country{Norway}}}


\abstract{As social robots increasingly enter public environments, their acceptance depends not only on technical reliability but also on ethical integrity, accessibility, and user trust. This paper reports on a pilot deployment of an ARI social robot functioning as a university receptionist, designed in alignment with the SecuRoPS framework for secure and ethical social robot deployment. Thirty-five students and staff engaged with the robot and provided structured feedback on safety, privacy, usability, accessibility, and transparency. Results show generally positive perceptions of physical safety, data protection, and ethical behaviour, while revealing persistent challenges around accessibility, inclusivity, and dynamic engagement.

Beyond these empirical findings, the study contributes in two further ways. First, it demonstrates how theoretical frameworks for ethical and secure design can be operationalised in practice through real-world deployment and end-user evaluation. Second, it provides a publicly available GitHub repository containing reusable templates for ARI robot applications, lowering barriers for beginners and supporting reproducibility in human–robot interaction research. By situating user perspectives at the centre of evaluation while offering practical resources for researchers, this work advances ongoing debates in AI and Society and contributes to the development of reliable intelligent environments that are trustworthy, inclusive, and ethically responsible in public spaces.}

%%================================%%
%% Sample for structured abstract %%
%%================================%%




\maketitle

\section{INTRODUCTION}\label{sec1}
Social robots are no longer confined to laboratories or exhibitions; they are increasingly deployed in public spaces such as shopping malls \cite{khanSystematicReviewSocial2024}, airports \cite{linAirportRobotsAutomation2023}, hospitals \cite{ragnoApplicationSocialRobots2023a}, and educational institutions \cite{lampropoulosSocialRobotsEducation2025}. In these environments, robots are expected to serve as assistants, guides, and information providers, complementing existing human service and care infrastructures \cite{rodsetholSocialRobotsPublic2022}. While these applications highlight the promise of social robotics, they also raise profound societal concerns. Issues of privacy, trust, accessibility, and ethical behaviour become especially pressing in uncontrolled, high-traffic environments where diverse users interact with the technology \cite{orumaSecurityAspectsSocial2023}.

Scholarly and industrial frameworks have begun to address the technical and ethical challenges of social robotics, yet most evaluations remain either conceptual or expert-driven \cite{orumaEnhancingSecurityPrivacy2026,  blaurockTransdisciplinaryReviewFramework2022, callariEthicalFrameworkHumanrobot2024, mollerNISTCybersecurityFramework2023}. Frameworks such as IEEE’s Ethically Aligned Design provide guiding principles but lack practical implementation strategies \cite{shahriariIEEEStandardReview2017}, while security-oriented models such as NIST or MITRE focus narrowly on digital vulnerabilities without addressing the broader human and societal dimensions of embodied robots in public spaces. As a result, a gap persists between theoretical guidance and end-user experience in real-world deployments \cite{nistNISTCybersecurityFramework2023, mitreMITREATTCK2025}.

The SecuRoPS (\underline{Secu}rity Framework for Social \underline{Ro}bots in \underline{P}ublic \underline{S}paces) framework was recently proposed to bridge this gap by offering a lifecycle approach that integrates security, safety, ethical compliance, and user-centred design \cite{orumaEnhancingSecurityPrivacy2026}. While its conceptual validity has been established in prior studies \cite{orumaUsercentredSecurityFramework2023,orumaArchitecturalViewsSocial2025}, little is known about how end users perceive robots developed under its guidance. Do the framework’s principles resonate with everyday users? Do they enhance perceptions of trust, safety, and inclusivity?

This paper addresses these questions through a pilot deployment of an ARI robot functioning as a receptionist at Østfold University College. The study engaged thirty-five participants, who interacted with the robot and provided structured feedback on their perceptions of safety, privacy, usability, accessibility, and transparency. By analysing these experiences, the study offers empirical insights into how theoretical frameworks translate into practice.

The contribution of this paper is fourfold. 
\begin{enumerate}
    \item \textbf{Empirical Validation in Public Space} – It presents one of the first in-situ, end-user validations of a lifecycle-based framework for secure and ethical robot deployment, conducted in a real university reception environment.
    \item \textbf{Societal Insights into Trust, Ethics, and Accessibility} – It provides rich qualitative and quantitative evidence showing how users perceive safety, privacy, inclusivity, and transparency when interacting with a public-facing robot, highlighting gaps in accessibility and the role of societal narratives in shaping trust.
    \item \textbf{Bridging Frameworks and Practice} – It demonstrates how theoretical principles of ethical and secure design can be operationalised in practice, offering lessons that extend beyond compliance to anticipatory and inclusive design strategies.
    \item \textbf{Open-Source Technical Contribution} – It delivers a publicly available GitHub repository containing reusable templates and implementation resources for the ARI platform, lowering barriers for beginners in robotics research and enabling reproducibility across institutions.
\end{enumerate}

The structure of the paper is as follows: Section \ref{sec2} situates this study within related work on security, ethics, and inclusivity in social robotics, highlighting both technical frameworks and societal debates. Section \ref{sec3} outlines the methodology, including the research design, participant recruitment, system implementation, and ethical considerations guiding the pilot deployment. Section \ref{sec4} presents the results of the end-user study, combining quantitative assessments and qualitative insights into trust, safety, usability, accessibility, and transparency. Section \ref{sec5} discusses these findings in relation to responsible robotics and HRI, drawing attention to broader societal implications, limitations, and threats to validity, as well as comparing the study with established security frameworks. Finally, Section \ref{sec6} concludes the paper by summarising key contributions, emphasising the practical value of the open-source repository, and identifying future directions for the responsible deployment of social robots in public spaces.

\section{RELATED WORK}\label{sec2}
\subsection{Social Robots in Public Spaces: Contexts and Functions}
Social robots have transitioned from controlled laboratories to everyday public settings, including educational institutions \cite{lampropoulosSocialRobotsEducation2025}, healthcare facilities \cite{ragnoApplicationSocialRobots2023a}, transportation hubs \cite{deveciAssessingAlternativesIncluding2023a}, and retail environments \cite{subero-navarroProposalModelingSocial2022}. In universities and classrooms, work has examined acceptance, usability, and pedagogical value, highlighting the importance of interaction design and multi-modality for engagement \cite{contiUseSocialRobots2025, lampropoulosSocialRobotsEducation2025}. In hospitals and care environments, reviews emphasize technical and organizational requirements alongside ethical constraints \cite{ragnoApplicationSocialRobots2023a,ragnoApplicationSocialRobots2023a}. Airport and retail deployments further underscore the need for robust interaction flows and clear role expectations in crowded, time-pressured spaces \cite{linAirportRobotsAutomation2023, khanSystematicReviewSocial2024,subero-navarroProposalModelingSocial2022}. Across these domains, a recurring theme is that deployment context, with its norms, crowd dynamics, and user goals, shapes perceptions of utility, safety, and trust.

\subsection{Ethics, Privacy, and Responsible Social Robotics}
Contemporary scholarship stresses that acceptance depends not only on technical reliability but also on ethical integrity, transparency, and inclusivity \cite{torrasEthicsSocialRobotics2024, stock-homburgEthicalConsiderationsCustomer2025, hungEthicalConsiderationsUse2025}. Practical guidance has begun to emerge: Callari et al. propose expert-informed ethical frameworks for human–robot collaboration \cite{callanderNavigatingHumanRobotInteraction2024}, while the IEEE Ethically Aligned Design principles foreground human well-being and transparency in AI systems \cite{shahriariIEEEStandardReview2017}. Yet these resources often remain high-level, with limited evidence on how to operationalize principles in situated deployments. Recent work on data minimization offers concrete tactics to reduce unnecessary data collection and processing \cite{staabPrinciplePracticeVertical2024}, aligning with GDPR’s spirit, but again, implementations in embodied public-space robots are still under-documented.

\subsection{Cybersecurity Frameworks vs. Embodied, Public-Facing Robots}
Mainstream cybersecurity frameworks, such as the NIST Cybersecurity Framework (CSF), MITRE ATT\&CK, and the Cyber Kill Chain, offer robust models for risk assessment, threat classification, and incident response in digital infrastructures \cite{nistNISTCybersecurityFramework2023, mitreMITREATTCK2025, lockheedmartinCyberKillChain2022, mollerNISTCybersecurityFramework2023}. They provide well-established taxonomies of adversarial behavior, tools for vulnerability management, and structured incident-handling processes that are highly effective for enterprise IT systems.

However, these frameworks were not designed for embodied, interactive agents operating in open public environments. Social robots introduce risks that extend beyond traditional digital threats, encompassing physical safety, bystander privacy, accessibility, inclusivity, and broader human factors integral to HRI. For example, an embodied robot must simultaneously protect against network intrusion and ensure that its physical presence does not cause harm or exclusion. These dimensions of reliability and trustworthiness are largely absent from conventional IT security models.

Practical secure-by-design measures in robotics therefore require additional layers of consideration, including runtime containment (e.g., containerized execution), hardened communication channels, and supply-chain integrity. Recent evidence on vulnerabilities in container ecosystems demonstrates both the opportunities and risks of such approaches \cite{shiDrDockerLargeScale2025}. Bridging IT-centric cybersecurity with sociotechnical safety thus remains an open challenge for public-space robotics.

The SecuRoPS framework responds to this gap by extending security concerns into a lifecycle perspective that explicitly incorporates resilience, ethical safeguards, inclusivity, and user trust. By integrating technical robustness with human-centered design principles, SecuRoPS contributes a more holistic model of reliable deployment—one that is sensitive not only to digital reliability but also to the embodied, interactive, and societal dimensions of robots in public environments.

\subsection{Trust, Acceptance, and Transparency in HRI}
A large body of HRI research examines the antecedents of trust (appearance, voice, motion, transparency) and acceptance (perceived usefulness, ease of use, social presence). Studies in customer service and hospitality highlight the role of rapport, interaction naturalness, and expectation management \cite{songUnderstandingTrustRapport2024, dingCustomerAcceptanceFrontline2024}. Work on mind perception and implicit associations shows how subtle design cues shape whether users ascribe agency or benevolence to robots \cite{pekcetinRealWorldImplicitAssociation2024}. Philosophical and conceptual critiques warn that trust without trustworthiness is possible if surface cues are decoupled from robust safeguards, making verifiable transparency essential \cite{massaguergomezShouldWeTrust2025}. In practice, research consistently shows that visible data practices, plain-language explanations, opt-in choices, and clear signals of GDPR compliance improve user confidence \cite{callanderNavigatingHumanRobotInteraction2024}.

\subsection{Accessibility and Inclusivity in Public Deployments}
Accessibility is often underrepresented in evaluations, despite being a central component of equitable public services. Educational deployments report strong usability when interfaces are clear and multimodal \cite{contiUseSocialRobots2025}, yet real-world cases reveal persistent barriers: screen height for wheelchair users, insufficient font or contrast, lack of speech input or sign language, and limited support for non-dominant languages \cite{barfieldDesigningSocialRobots2023}. As robots transition into public infrastructures (libraries, hospitals, transport hubs), accessibility must evolve from a compliance checklist to an anticipatory design principle that proactively accommodates diverse abilities and contexts.

\subsection{From Threat Landscape to Lifecycle Governance: SecuRoPS}
Survey and mapping studies have characterized the threat landscape for social robots in public spaces, documenting attack surfaces, privacy risks, and organizational challenges \cite{orumaSystematicReviewSocial2022}. Building on this evidence, SecuRoPS was proposed as a lifecycle governance framework that integrates risk assessment, ethical and legal scrutiny, usability, and stakeholder engagement across phases from analysis to retirement \cite{orumaUsercentredSecurityFramework2023}. Subsequent architectural work elaborated business, system, and security views tailored to public-space deployments \cite{orumaArchitecturalViewsSocial2025}. Yet, despite conceptual maturation, end-user validation of such frameworks in real-world conditions has been scarce.

\subsection{Reproducibility, Tooling, and Entry Barriers}
HRI deployments often rely on proprietary SDKs and complex integration stacks (web front-end + ROS back-end), creating entry barriers for new researchers and institutions. Educational texts and tutorials support onboarding (e.g., ROS introductions) \cite{josephRobotOperatingSystem2022}, but reusable, open templates specific to public-space deployments remain limited. Open-source, well-documented exemplars are vital to reproducibility and to translating ethical and security principles into deployable artefacts that newcomers can adapt.

\subsection{Positioning This Study}
Against this backdrop, the present work contributes along two axes. First, it provides in-situ, end-user validation of selected SecuRoPS phases in a public educational setting, connecting framework ideals to user perceptions of safety, privacy, usability, accessibility, and transparency. Second, it delivers a publicly available GitHub repository with reusable ARI templates and data-handling modules, thereby lowering entry barriers for robotics beginners and supporting reproducibility in HRI. In doing so, this study addresses the documented gaps between high-level guidance and operational practice, as well as between lab-based evaluations and the complexities of public-space interaction.

\begin{figure}[h]
    \centering
    \includegraphics[width=0.99\linewidth]{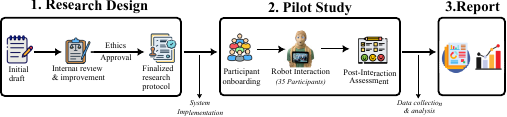}
    \caption{Overview of the research methodology followed in the study}
    \label{fig002}
\end{figure}

\section{METHODOLOGY}\label{sec3}
%\subsection{Research Design}
This study employed a qualitative pilot design to explore how end users perceive a social robot deployed in a real-world educational context. An overview of the research methodology followed in this study is illustrated in Figure \ref{fig002}, which outlines the sequential steps from ethical approval to data reporting. Selected phases of the SecuRoPS framework for secure and ethical social robot deployment were operationalised \cite{orumaEnhancingSecurityPrivacy2026}, with a focus on user perceptions of safety, privacy, usability, accessibility, and ethical transparency. By situating the robot in a university reception area, the study enabled participants to engage with it in a realistic public setting rather than a controlled laboratory environment.

%\subsection{Ethical Considerations}
The research protocol was reviewed and approved by the Norwegian Agency for Shared Services in Education and Research (Sikt\footnote{https://sikt.no/}, ref. 328081). Participants were fully informed about the study, including the voluntary nature of participation, the option to withdraw at any time, and the intended use of their data. In line with GDPR principles, only minimal personal data were collected (optional name and email), and all robot cameras were physically masked to prevent inadvertent recording \cite{radley-gardnerFundamentalTextsEuropean2020}. These measures ensured that transparency and respect for privacy were embedded into the design and conduct of the study.

%\subsection{Participants}
\begin{figure}[h]
    \centering
    \includegraphics[width=0.9\linewidth]{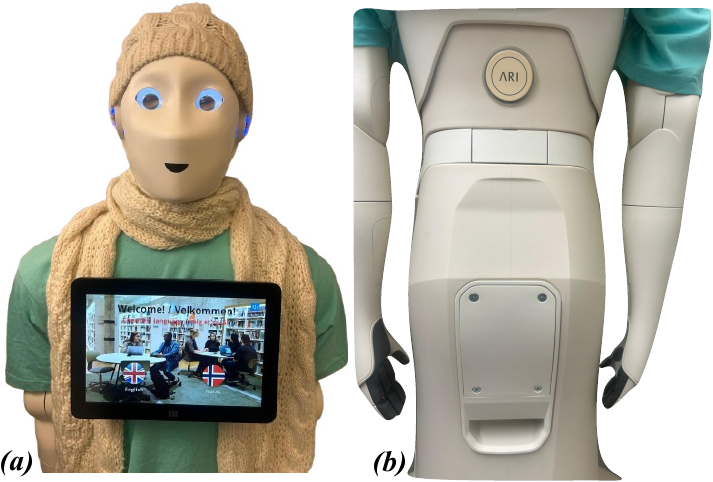}
    \caption{Branded Ari Social Robot utilized in this study (a) Front view, (b) Back view}
    \label{fig003}
\end{figure}

%\subsection{System Implementation}
The study was conducted using an ARI social robot programmed to act as a university receptionist. Figure \ref{fig003}  presents the front and back views of the Ari Social Robot used for this study. The robot delivered bilingual (English and Norwegian) information on study programmes and campus life through web-based slides, videos, and interactive presentations. A custom virtual keyboard allowed participants to provide their name and email address, optionally, testing perceptions of transparency in data collection. All entries were stored locally on the robot and deleted upon request.

The software was developed using standard web technologies (HTML, CSS, and JavaScript) with a ROS backend, and is made publicly available through a dedicated GitHub repository \cite{orumaARIOstfoldUniversityCollegeRobotReceptionist2025}. This repository contains reusable templates and implementation examples, designed to lower barriers for researchers and educators who are new to proprietary robotic platforms such as ARI. Technical safeguards, including VPN-enabled communication, restricted execution environments, and firewall protections, were applied in accordance with security-by-design principles \cite{orumaArchitecturalViewsSocial2025}. These measures not only addressed security and privacy, but also enhanced the system’s reliability by preventing unplanned behaviours, reducing error propagation, and ensuring consistent operation throughout the deployment.

Thirty-five participants were recruited from Østfold University College, comprising both staff (48.6\%) and students (45.7\%) across multiple faculties. Recruitment was voluntary and open, resulting in diversity in age, gender, academic background, and prior exposure to social robots. The sample size is consistent with exploratory HRI studies \cite{pekcetinRealWorldImplicitAssociation2024}, providing both breadth of perspectives and depth of qualitative feedback appropriate for pilot validation.

%\subsection{Procedure}
The study followed four sequential phases: \\ 
\noindent \faBattery[1] \textbf{ Phase 1: Informed Consent:} Participants were initially greeted by the researcher and given access to the Participant Information Sheet through a digital QR code. They provided informed consent digitally via Nettskjema, ensuring clear documentation of their voluntary participation and understanding of the study (See Appendix \ref{secA}). 

\noindent \faBattery[2] \textbf{ Phase 2: Demographic Information Collection:} Following informed consent, participants completed a demographic information form provided through a second QR code linking to another secure Nettskjema form. This demographic data included age, gender, faculty affiliation, and previous experiences with social robots, which served as contextual background information for analysis (See Appendix \ref{secB}). 

\noindent \faBattery[3] \textbf{ Phase 3: Interaction with ARI Social Robot:} Participants interacted with the ARI social robot, which was programmed to function as a receptionist at Østfold University College. 
The robot provided visual and verbal information in both English and Norwegian through: 
\begin{itemize}
    \item Web page presentations highlighting university study programs.
    \item Interactive picture slideshows showcasing campus life.
    \item Video presentations about the college.
    \item PowerPoint presentations detailing specific academic offerings.
\end{itemize}

Participants were invited to use a custom-designed virtual keyboard integrated within the robot's interface to provide their names and email addresses, optionally. This exercise aimed to evaluate the robot's transparency in data collection processes. Participants engaged with the robot following the research guidelines outlined in Appendix \ref{secC}. 

\noindent \faBattery[4] \textbf{ Phase 4: Post-Interaction Assessment:} After interacting with ARI, participants completed a structured feedback survey through Nettskjema, addressing their perceptions related to physical safety, data privacy, protection from physical tampering, cybersecurity vulnerabilities, interface usability, accessibility accommodations, ethical behaviour and transparency in data collection, and compliance with privacy laws such as GDPR (See Appendix \ref{secD}). The feedback gathered provided insights into the effectiveness of the SecuRoPS framework in practical, public interaction scenarios.

%\subsection{Data Management and Analysis}
All consent forms, demographic data, and survey responses were stored securely via Nettskjema and Østfold University’s Office 365 system. Optional personal information (name/email) collected during interaction was stored locally on the robot and excluded from analysis. To ensure confidentiality, all data were anonymised before processing. Quantitative responses were analysed descriptively, while qualitative feedback was examined thematically to identify patterns across trust, usability, accessibility, and transparency.

\section{RESULTS}\label{sec4}
\subsection{Participant Demographics}
The study engaged 35 participants, comprising a balanced gender distribution (51\% male, 49\% female) and a wide age range. Most were university staff (48.6\%) or students (45.7\%), with representation across faculties including computer science, teacher education, and health. A majority (62.9\%) had never interacted with a social robot before, and 68.6\% reported no prior use of robots in education. This diversity provided an opportunity to capture perspectives from both novice and moderately experienced users.
\begin{table}[h]
\centering
\caption{Participant responses (\%) to key post-interaction assessment questions.}
 \label{tab1}
\begin{tabular}{lcc} 
\toprule
\multicolumn{1}{c}{\textbf{Evaluation   Criteria}} & \multicolumn{1}{c}{\textbf{Positive (\%)}} & \multicolumn{1}{c}{\textbf{Negative (\%)}} \\
\midrule
Physical   Safety                         & 80.0                                & 20.0                                \\
Privacy   Concern                         & 85.3                              & 14.7                              \\
Physical   Security                       & 89.7                              & 10.3                              \\
Cybersecurity                             & 84.8                              & 15.2                              \\
Interface   Usability                     & 79.4                              & 20.6                              \\
Accessibility                             & 60.6                              & 39.4                              \\
Ethical   Behavior \& Transparency        & 82.4                              & 17.6                              \\
GDPR   Compliance                         & 90.0                                & 10.0             \\
\bottomrule
\end{tabular}
\end{table}
\subsection{Quantitative Findings}
\begin{figure}[h]
    \centering
    \includegraphics[width=0.99\linewidth]{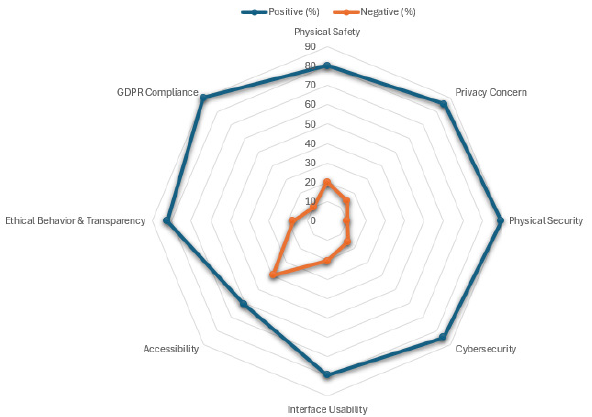}
    \caption{Radar chart showing positive and negative participant responses (\%) across key post-interaction assessment criteria}
    \label{fig001}
\end{figure}

As shown in Table \ref{tab1} and Figure \ref{fig001}, participants generally evaluated the robot positively across core dimensions. High levels of trust were expressed regarding physical safety (80\%), data privacy (85.3\%), and GDPR compliance (90\%). Similarly, physical security (89.7\%) and cybersecurity protections (84.8\%) were perceived favourably, suggesting that safeguards and transparency measures were effective in building confidence.

Usability (79.4\%) and ethical behaviour (82.4\%) were also endorsed, although accessibility received the lowest rating (60.6\%), highlighting a persistent challenge in designing inclusive systems.

\subsection{Qualitative Insights}
Open-ended feedback enriched these results, revealing three main themes:
\begin{itemize}
    \item \textit{Trust and Friendliness:} Participants consistently described the robot as \textit{“safe,” “friendly,”} and \textit{“easy to use,”} with many appreciating its non-threatening design and dual-language support. These perceptions reflect the importance of appearance, multimodality, and transparency in fostering trust \cite{changCrossmodalInteractionsHumanRobot2025,torrasEthicsSocialRobotics2024,vitaleBeMoreTransparent2018}.
    \item \textit{Usability and Interaction Flow:} While most participants found the interface intuitive, suggestions included clearer navigation cues, a larger on-screen keyboard, and a more natural voice. Some recommend adding motion or voice control to make interactions more dynamic and engaging \cite{contiUseSocialRobots2025}.
    \item \textit{Accessibility and Inclusivity:} Accessibility concerns emerged strongly. Participants noted that the screen height was challenging for wheelchair users and that features such as speech input or sign language support were lacking \cite{dehnert2024ability,desailleImprovingInclusivityRobotics2022}. These insights underline the need for more inclusive design beyond compliance-based approaches.
    \item \textit{Transparency in Data Practices:} While GDPR compliance was recognised, several participants suggested making privacy assurances more explicit during interaction (e.g., the robot verbally stating its compliance). This highlights the role of visible transparency in building user trust \cite{songUnderstandingTrustRapport2024,callanderNavigatingHumanRobotInteraction2024}.
\end{itemize}

Overall, the findings suggest that while the robot was broadly trusted and well-received, accessibility and dynamic interaction remain critical areas for improvement in future deployments.

\subsection{Mapping Pilot Study Activities to the SecuRoPS Framework}
The pilot study conducted at Østfold University College 
aligns with several phases and functional components of the SecuRoPS framework. Table~\ref{tab002} provides a summary of how the study activities correspond to the framework’s 14 phases and its 7 core functions \cite{orumaEnhancingSecurityPrivacy2026}.

\begin{table}[h!]
\centering
\caption{Mapping of Pilot Study Activities to the SecuRoPS Framework Phases and Functions}
\footnotesize
\begin{tabular}{p{4cm}p{8cm}}
\toprule
\textbf{SecuRoPS Phases} & \textbf{Pilot Study Activity} \\
\midrule
Business Needs Assessment & Identified the need for evaluating secure, ethical, and inclusive deployment of a social robot in an educational setting. \\
\midrule
Evaluation of SRPS Application Context & Østfold University College
was selected as a realistic and relevant public space for deployment. \\
\midrule
Robot Type Selection & ARI robot chosen based on its capabilities, compatibility with Docker, and development environment constraints. \\
\midrule
Stakeholder Engagement and Dialogue & Collaboration with the university communication department and ethics committee (Sikt) to ensure alignment and compliance. \\
\midrule
Feasibility and Impact Analysis & Considered accessibility needs, user privacy concerns, and potential institutional benefits. \\
\midrule
Requirement Specification & Developed application requirements for multilingual support, audio/visual interaction, secure data handling, and usability. \\
\midrule
Risk Assessment \& Threat Modelling & Security risks mitigated through covered ports, VPN/firewall, and stationary operation. Data minimization enforced through design. \\
\midrule
Proven Methodology-Driven Design & Followed PAL Robotics SDK implementation methods and guidelines for structured application design. \\
\midrule
Implementation of Security, Safety, and User-Centeredness Measures & Built secure HTML/CSS/JS-based UI; implemented overlay navigation, visual/audio aids, and GDPR-aligned data collection process. \\
\midrule
User Experience, Usability, and Security Testing & Post-interaction surveys and direct observation provided feedback on trust, safety, and accessibility. \\
\midrule
Ethical, Legal, and Regulatory Scrutiny & Received formal approval from Sikt; data handling compliant with Norwegian data protection laws. \\
\midrule
Strategic Deployment & Robot deployed in a public-facing location at the university with pre-programmed interactive content. \\
\midrule
Continuous Monitoring and Iterative Improvement & On-site observations and usability feedback recorded for future design improvements. \\
\midrule
Retrospective and Lessons Learned & Reflections incorporated into recommendations and design templates published with the study. \\
\bottomrule
\end{tabular}
 \label{tab002}
\end{table}

This mapping demonstrates that the pilot study addressed activities associated with all 14 phases of the framework.

\section{DISCUSSION}\label{sec5}
This study provides one of the first empirical validations of a structured framework for secure and ethical deployment of social robots in public spaces, focusing on end-user perspectives in a university setting \cite{orumaEnhancingSecurityPrivacy2026}. Overall, participants perceived the robot positively in terms of physical safety, data privacy, and ethical transparency, while also identifying gaps in accessibility and interactivity. These findings extend debates in HRI and responsible AI by demonstrating how real-world users respond to robots designed with explicit attention to ethical, security, and inclusivity concerns \cite{callanderNavigatingHumanRobotInteraction2024}.

\subsection{Trust, Safety, and Ethical Transparency}
Trust is a cornerstone of social robot acceptance, particularly in uncontrolled public spaces \cite{massaguergomezShouldWeTrust2025}. In this study, participants consistently rated the robot as physically safe (80\%) and ethically transparent (82.4\%). The robot’s friendly appearance and bilingual communication contributed to this trust, aligning with prior evidence that design cues and multimodal interfaces reduce perceived risk \cite{changCrossmodalInteractionsHumanRobot2025,caoInvestigatingRoleMultimodal2023}. Importantly, ethical transparency was not only about the absence of harm but also about clarity in data practices. Participants valued GDPR compliance but wanted this to be made more visible during interaction—for example, by the robot explicitly stating its compliance. This highlights the importance of real-time, user-facing transparency as a trust-building mechanism, complementing legal compliance \cite{aroraVirtuousIntegrativeSocial2025}.

\subsection{Accessibility and Inclusivity}
Accessibility emerged as the weakest dimension, with only 60.6\% of participants affirming inclusivity. Users pointed out limitations for wheelchair users, visually impaired individuals, and those requiring alternative communication modes (e.g., speech input or sign language). This underscores that compliance-based accessibility (e.g., readable text, bilingual support) is not enough; instead, inclusivity requires proactive design choices that anticipate diverse needs. In societal terms, inadequate accessibility risks reinforcing digital and social exclusion, especially as robots enter public infrastructures such as hospitals, libraries, and transport hubs. The findings stress that accessibility should not be a “phase” of design but a core principle of responsible social robotics \cite{dehnert2024ability,desailleImprovingInclusivityRobotics2022}.

\subsection{Transparency, Privacy, and GDPR in Practice}
The high endorsement of GDPR compliance (90\%) shows that participants appreciated data minimisation (e.g., no camera use, optional email collection). However, the qualitative feedback reveals that compliance alone is insufficient for building trust. Users sought visible and explicit communication about privacy safeguards. This suggests that future robots should incorporate privacy-by-interaction, where assurances are actively conveyed during dialogue rather than buried in protocols \cite{songUnderstandingTrustRapport2024,callanderNavigatingHumanRobotInteraction2024}. In a societal context where anxieties about surveillance are strong, visible privacy practices are critical to public legitimacy of AI-driven technologies.

\subsection{User Expectations, Bias, and Media Narratives}
Despite never interacting with robots before, many participants voiced concerns about surveillance and misuse of data. This reflects a societal bias shaped more by cultural narratives (films, media, public discourse) than by actual experience. Such biases can both hinder acceptance and highlight the importance of cultural framing in robot deployment \cite{dorafshanianDataCollectionSafeguarding2024}. Addressing these expectations requires not just technical safeguards but also public engagement strategies that demystify robots and correct misconceptions.

\subsection{Interaction Dynamics and Attention Economy}
Another emergent finding was user impatience. Many participants skipped through content, ignoring features like countdown timers. This behaviour reflects the attention economy of public spaces, where users may not engage deeply with robots unless interactions are concise, personalised, and immediately relevant \cite{damianoHomesHumanRobot2021,arduengoRobotEconomyHere2021}. For designers, this means prioritising critical information early in the interaction flow and considering adaptive systems that respond to user impatience. This finding connects to broader debates on human–AI cohabitation in public infrastructures, where time constraints and competing demands shape engagement.

\subsection{Contribution to Responsible Robotics and HRI}
Compared with many HRI studies that rely on laboratory simulations or expert evaluations, this research demonstrates the unique value of in-situ, end-user validation. By deploying the robot in a real university environment, the study captured authentic behaviours, expectations, and biases that cannot be replicated in controlled settings.

A notable contribution of this work is the creation of a publicly available GitHub repository, containing reusable templates and implementation resources for the ARI platform \cite{orumaARIOstfoldUniversityCollegeRobotReceptionist2025}. This resource lowers the barrier of entry for researchers and educators new to proprietary robotic systems, supporting reproducibility and adaptation across contexts. By combining empirical findings with practical, open-source tools, the study exemplifies how responsible robotics research can connect societal insights with technical enablers, thereby advancing both theory and practice.

\subsection{Comparison with Established Security Frameworks}
Traditional frameworks such as NIST, MITRE ATT\&CK, and the Cyber Kill Chain provide robust models for cybersecurity but are largely limited to digital infrastructures \cite{nistNISTCybersecurityFramework2023,mitreMITREATTCK2025,lockheedmartinCyberKillChain2022}. They do not address embodied risks such as physical safety, inclusivity, or public trust. The SecuRoPS framework extends these models by embedding human-centred, lifecycle-based considerations into deployment. Unlike traditional frameworks, SecuRoPS includes stakeholder dialogue, user testing, and iterative learning as core phases \cite{orumaEnhancingSecurityPrivacy2026}. This positions it as a more detailed approach to sociotechnical security, bridging technical protection with societal acceptance.

\subsection{Reliability and Lessons Learned}
A central concern in the deployment of intelligent environments is not only whether systems function as intended, but whether they do so reliably across varied conditions and user groups. Reliability in HRI has been broadly understood as the ability of a system to deliver consistent performance despite environmental variability and heterogeneous user needs \cite{mavrogiannisCoreChallengesSocial2023, ragnoApplicationSocialRobots2023a}. This study highlights several reliability lessons relevant to social robots in public spaces.

On the technical side, the system demonstrated dependable performance throughout the pilot. Measures such as containerised execution within the PAL Robotics SDK, firewall protections, and VPN-enabled communication ensured that the robot operated securely and without technical interruptions. Local storage of optional user data prevented transmission errors and enhanced resilience in handling sensitive information. These safeguards exemplify reliability-by-design, where robustness and fault containment are embedded into the system architecture \cite{shiDrDockerLargeScale2025, josephRobotOperatingSystem2022}.

At the same time, reliability gaps emerged from the interaction context. The stationary configuration of the robot limited its capacity to adapt dynamically, and some participants perceived it as \textit{“too static.”} Similarly, accessibility challenges, such as screen height for wheelchair users and lack of alternative input modalities, demonstrated that reliable performance was not uniformly experienced across diverse participants. From a system reliability perspective, these are not failures of functionality but failures of consistency, where the system delivers unevenly depending on user needs or situational constraints \cite{contiUseSocialRobots2025, stock-homburgEthicalConsiderationsCustomer2025}.

Another lesson relates to interaction pacing. User impatience, reflected in skipping through content or ignoring timers, indicates that reliability in public environments must account for the attention economy. Even if the system operates without fault, its perceived reliability diminishes if it cannot sustain engagement under real-world time pressures. Designing adaptable interaction flows that prioritise critical content early and adjust to user impatience will be essential for reliable public deployments \cite{songUnderstandingTrustRapport2024, pekcetinRealWorldImplicitAssociation2024}.

Together, these findings illustrate that reliability in intelligent environments is multifaceted: it includes technical robustness, but also equitable consistency across diverse users, resilience under varying interaction conditions, and adaptability to the sociocultural narratives shaping trust. These lessons extend beyond a single case study, offering transferable insights for the design of reliable, trustworthy, and socially embedded robots in public spaces \cite{torrasEthicsSocialRobotics2024, callanderNavigatingHumanRobotInteraction2024}.

\subsection{Limitations}
This study provides valuable insights into user perceptions of a social robot in a public educational setting, yet several limitations must be acknowledged. Recognising these limitations helps clarify the scope of the findings and guide future HRI research.
\begin{enumerate}
    \item \textit{Contextual Scope:} The study was conducted in a single institution (Østfold University College), which limits the extent to which findings can be generalised to other contexts such as healthcare, transportation, or retail. While the inclusion of both staff and students across multiple faculties introduced diversity of perspectives, the institutional setting may have shaped user expectations (e.g., familiarity with university information may have reduced novelty). This limitation was partially addressed by recruiting participants from different roles, age groups, and backgrounds. However, future studies should extend deployment to multiple public environments to capture context-dependent variations in trust, accessibility, and inclusivity.
    \item \textit{Sample Size and Representativeness:} With 35 participants, the sample is modest compared to large-scale surveys but consistent with exploratory HRI pilot studies \cite{pekcetinRealWorldImplicitAssociation2024}. The study prioritised depth of qualitative feedback over statistical generalisability. To address representativeness, we ensured balance in gender and age distribution, and included participants with and without prior experience of robots. Nonetheless, larger and more heterogeneous samples are needed in future work to confirm patterns and explore demographic effects in greater depth.
    \item \textit{Interaction Duration and Scope:} Participant interactions were short-term and limited to a predefined set of tasks (information delivery, optional data entry). This design enabled focus on transparency and usability but restricted insights into long-term acceptance, adaptation, or behaviour over repeated interactions. Measures taken to mitigate this limitation included combining structured surveys with open-ended feedback to capture both immediate impressions and nuanced reflections. Future studies should incorporate longitudinal deployments, allowing participants to interact over days or weeks to better understand sustained engagement, trust evolution, and potential novelty effects.
    \item \textit{Robot Capabilities:} The ARI robot was configured as a stationary unit, without autonomous navigation or expressive gesturing. While this reduced safety risks in a crowded space, it also limited the richness of interaction. Some participants explicitly noted the robot felt \textit{“too static.”} This limitation was acknowledged transparently, and participant feedback was collected to inform improvements (e.g., mobility, voice input, naturalistic behaviours). Future HRI studies should deploy robots with a broader range of capabilities to test how mobility, adaptivity, and expressivity influence perceptions of trust and inclusivity.
    \item \textit{Task and Content Familiarity:} Since the robot presented university-specific information, some participants—particularly staff—were already familiar with the content. This may have contributed to impatience or disengagement. To reduce this bias, we recruited both staff and students from different faculties, including individuals less exposed to the university’s communication materials. Nonetheless, future studies should test robots with novel or personalised content to evaluate how information relevance shapes engagement and acceptance.
    \item \textit{Self-Reported Measures:} Finally, reliance on self-reported survey data introduces risks of social desirability or bias. While anonymity and voluntary participation reduced these pressures, responses may still not perfectly reflect actual perceptions or behaviours. To mitigate this, we triangulated self-reported data with direct observations and qualitative comments, strengthening interpretive validity. Future HRI studies should incorporate behavioural measures (e.g., interaction logs, dwell times, or physiological responses) to complement subjective feedback.

\end{enumerate}

\subsection{Threats to Validity and Mitigation Measures}
Like all empirical studies, this work is subject to potential threats to validity. Anticipating and addressing these threats is essential for interpreting the findings responsibly \cite{seamanQualitativeResearchMethods2025,sjobergConstructValiditySoftware2023}. Below, we outline four categories of validity concerns, provide examples from this study, and describe mitigation measures.
\begin{itemize}
    \item \textit{Internal Validity:}
Internal validity may have been affected by participant bias, particularly social desirability in responses. For example, some participants may have felt pressure to provide positive feedback because the robot was deployed in their own institution, or because they were interacting in the presence of a researcher. To mitigate this, all survey responses were collected anonymously via Nettskjema, ensuring that individuals could express negative feedback without attribution. Participants were also reminded that the study was non-evaluative and that critical opinions were valuable.
    \item \textit{Construct Validity:}
A threat to construct validity arises when measurement tools fail to capture the intended phenomena. In this study, the structured survey was aligned with dimensions of the SecuRoPS framework (e.g., safety, accessibility, transparency). However, these abstract concepts may have been interpreted differently by participants. For instance, some may have equated “cybersecurity” with data privacy rather than network protection. To address this, the survey questions were phrased in plain language and accompanied by short clarifications (e.g., distinguishing between “physical safety” and “cybersecurity vulnerabilities”). In addition, we triangulated responses with qualitative feedback, enabling us to cross-check whether numerical ratings aligned with participants’ explanations in open-ended comments.
    \item \textit{External Validity:}
The generalisability of the findings is limited because the study was conducted in a single university context with a modest sample size (n=35). Participant familiarity with the university’s information materials may have influenced engagement levels, as some rushed through content they already knew. This raises the possibility that results would differ in environments where users have no prior familiarity (e.g., airports or libraries). We mitigated this limitation by ensuring diversity within the sample (age, gender, faculty, and staff vs. student roles), providing a range of perspectives. Nevertheless, we caution that replication in different public settings and with larger, more heterogeneous samples is necessary to validate the broader applicability of these findings.
    \item \textit{Technical Validity:}
The technical configuration of the robot introduced additional constraints. ARI was deployed as a stationary unit without autonomous navigation or expressive gestures, features that could significantly affect perceptions of engagement and safety. For example, users who found the robot “too static” may have evaluated it less positively than they would a mobile or more interactive system. These limitations were transparently reported, and participant suggestions (e.g., adding movement or voice input) were documented as future design improvements. Furthermore, technical safeguards (such as firewall protections and local storage of personal data) were applied consistently, ensuring that findings on trust and transparency reflected real system behaviours rather than hypothetical assurances.
\end{itemize}

\section{CONCLUSION}\label{sec6}
This study presented one of the first in-situ validations of a lifecycle-based framework for the secure and ethical deployment of social robots in public spaces. By situating the ARI robot in a university reception context and engaging end users directly, the study demonstrated that social robots can be perceived as safe, trustworthy, and ethically transparent when designed with explicit attention to security, usability, and regulatory compliance. At the same time, it revealed important challenges around accessibility, inclusivity, and dynamic engagement—areas that remain critical for ensuring that social robots serve all members of society equitably.

Beyond its empirical and ethical contributions, this study also advances the broader agenda of reliable intelligent environments. The deployment demonstrated how technical safeguards such as containerised execution, VPN-enabled communication, and local data storage can provide robust and dependable system performance in real-world contexts. At the same time, the pilot revealed reliability gaps in accessibility and engagement, showing that consistency across diverse users and contexts is as important as fault-free operation. By coupling reliability with transparency, inclusivity, and security, the study contributes to a holistic understanding of what it means for intelligent systems to operate dependably in public spaces. The accompanying open-source repository further ensures that these lessons are reproducible and transferable, enabling researchers and practitioners to design and deploy social robots that are not only ethical and trustworthy, but also reliably integrated into everyday environments.

Importantly, the study also makes a practical contribution to the robotics community. A publicly available GitHub repository accompanies this research, offering reusable templates and implementation resources for the ARI platform. By lowering technical barriers to entry, the repository enables beginners and institutions with limited robotics expertise to replicate and extend the study, fostering reproducibility and wider participation in HRI research.

Taken together, these contributions bridge the gap between abstract ethical frameworks and lived public experience. They provide actionable insights for policymakers, designers, and roboticists seeking to develop trustworthy, inclusive, and socially responsible robots for public spaces. Future work should expand this line of inquiry by testing deployments across multiple domains, such as healthcare, libraries, or transport hubs, and by exploring long-term interactions to capture evolving patterns of trust and engagement. In this way, social robots can be responsibly integrated into society not only as technological artefacts but as public agents that embody fairness, inclusivity, and transparency.

\section*{Declarations}
\bmhead{Funding}
This research was funded by the Norwegian Research Council under the SecuRoPS project “User-centered Security Framework for Social Robots in Public Spaces” with project code 321324.
\bmhead{Conflict of Interest/Competing Interest}
The authors affirm that human research participants provided informed consent for publication of their feedback and comments.
\bmhead{Ethics Approval}
The study design and interview protocol for this research were reviewed and approved by the Norwegian Agency for Shared Services in Education and Research (Sikt) with reference code 328081.
\bmhead{Informed Consent}
All participants provided informed consent before their inclusion in the study. The consent process ensured that participants were fully aware of the study's aims, procedures, potential risks, and their right to withdraw at any time without any consequences.
\bmhead{Consent to Publish}
The authors affirm that human research participants provided informed consent for publication of their feedback and comments.
\bmhead{Data availability statement}
The system implementation code is available on GitHub \faLink { https://github.com/samsonoo/ARI---Ostfold-University-College-Robot-Receptionist}.
% Acknowledgements are not compulsory. Where included they should be brief. Grant or contribution numbers may be acknowledged.

% Please refer to Journal-level guidance for any specific requirements.

% \section*{Declarations}

% Some journals require declarations to be submitted in a standardised format. Please check the Instructions for Authors of the journal to which you are submitting to see if you need to complete this section. If yes, your manuscript must contain the following sections under the heading `Declarations':

% \begin{itemize}
% \item Conflict of interest/Competing interests (check journal-specific guidelines for which heading to use)
% \item Ethics approval and consent to participate
% \item Consent for publication
% \item Materials availability
% \end{itemize}

% \noindent
% If any of the sections are not relevant to your manuscript, please include the heading and write `Not applicable' for that section. 

% %%===================================================%%
% %% For presentation purpose, we have included        %%
% %% \bigskip command. Please ignore this.             %%
% %%===================================================%%
% \bigskip

% Springer journals and proceedings: \url{https://www.springer.com/gp/editorial-policies}

% Nature Portfolio journals: \url{https://www.nature.com/nature-research/editorial-policies}

% \textit{Scientific Reports}: \url{https://www.nature.com/srep/journal-policies/editorial-policies}

% BMC journals: \url{https://www.biomedcentral.com/getpublished/editorial-policies}
% \end{flushleft}
\newpage
\begin{appendices}
\section{Participant Information Sheet}\label{secA}
\textbf{\textit{Title:}} Real-World Validation of a Human-Centred Social Robot: A University Receptionist Case Study  \\ 
\textbf{\textit{Invitation to Participate}}\\ 
You are invited to participate in a research study to validate the SecuRoPS framework from the end-users' perspective. Your feedback will help us evaluate some phases of the framework, focusing on security, safety, usability, user experience, and ethical considerations of social robots in public spaces. \\ 
\textbf{\textit{About the Study}} \\ 
In this study, you will interact with a social robot that provides information about Østfold University College, 
its campuses, departments, and courses, as well as updates on local events and attractions in Halden. Following your interaction with the ARI social robot, you will be asked to provide feedback on your experience, helping us test the SecuRoPS framework.\\ 
\textbf{\textit{Your Role in the Study}} \\
You will interact with the robot and complete a feedback survey afterwards, evaluating its security, usability, ethical considerations, and user experience. \\ 
\textbf{\textit{Voluntary Participation}} \\ Your participation is entirely voluntary. You may withdraw at any point without penalty, and your decision will not affect your relationship with Østfold University College 
or any affiliated institutions. \\ 
\textbf{\textit{Duration of the Study}} \\
The total time commitment is approximately 20 minutes. \\ 
\textbf{\textit{Data Collection and Confidentiality}} \\
No personal data will be collected during your interaction with the robot unless you choose to provide it. The robot will not collect any data from users through its camera or other sensors while providing information. However, the robot will ask users for their names and email addresses. You will have the option to either accept or decline this request. At the end of the interaction, the robot will review the information collected during the session and ask for your consent to save or discard this data. Please note that names and email addresses are not considered sensitive information under the data protection laws in Norway.

This data collection process will be used to assess the transparency of the social robot during interactions with users. After the interaction, the only data collected for research purposes will be your responses in the feedback survey. All collected data will be anonymized and stored securely in Nettskjema and Microsoft Office OneDrive of Østfold University College. 
These data will be used exclusively for the purposes of this study. \\ 
\textbf{\textit{Benefits and Risks}}  \\ Your participation will advance knowledge in the emerging field of social robotics in public spaces. There are no significant risks associated with this study. \\ 
\textbf{\textit{Funding}} \\ This research is part of a PhD project partially funded by the Research Council of Norway under “User-centred Security Framework for Social Robots in Public Spaces (SecuRoPS)" with project code 321324. \\ 
\textbf{\textit{Further Information and Contact}} \\
Please get in touch with any of the following if you have any questions or need additional information.\\ 
\textit{PhD Researcher} : Samson Oruma, \textit{samsonoo@hiof.no} \\ 
\textit{Principal Investigator} – Prof. Dr. Ricardo Colomo-Palacios; \textit{ricardo.colomo-palacios@hiof.no} \\ 
\textit{Chief Information Security Officer} – CISO: Ted Magnus Sørlie; \textit{ted.m.sorlie@hiof.no}  \\

\noindent Thank you for considering this opportunity to shape the future of social robotics

\section{Participant Demographic Information}\label{secB}
\begin{enumerate}
    \item Age: \\  
    $\Box$ Under 18,  $\Box$ $15-25$,  $\Box$ $26-35$,  $\Box$ $36-45$,  $\Box$ $46-55$,  $\Box$ Over 55 
    \item Gender: \\  
    $\Box$ Male,  $\Box$ Female,  $\Box$ Non-binary,  $\Box$ Prefer not to say 
    \item Faculty/Unit: \\         
             $\Box$ Faculty of Health, Welfare and Organisation \\
             $\Box$ Faculty of Computer Science, Engineering and Economics \\
             $\Box$ Faculty of Teacher Education and Languages  \\
             $\Box$ Norwegian Theatre Academy \\
             $\Box$ The Norwegian National Centre for English and other Foreign Languages in Education \\
             $\Box$ Library \\
             $\Box$ Other (Please specify) \rule{9.2cm}{0.5pt}
       
        \vspace{0.2cm}
        \item Previous Experience with Social Robots (if any): \\
            $\Box$ None \\
            $\Box$ Some experience \\
            $\Box$ Extensive experience 
            
            \vspace{0.2cm}
        \item Have you interacted with robots for educational purposes before? \\  $\Box$	Yes, 		$\Box$  No 	
\end{enumerate}
\newpage
\section{Research Guideline}\label{secC}
\subsection*{Title:  Real-World Validation of a Human-Centred Social Robot: A University Receptionist Case Study}
\subsection*{Overview of Research}
This study evaluates the SecuRoPS framework from the perspective of end-users, focusing on the framework’s ability to address concerns related to security, usability, user experience, and ethical considerations in social robots deployed in public spaces.
\subsection*{Research Procedure}
\begin{enumerate}
    \item \textbf{Introduction:} 
        \begin{itemize}
            \item Greet the participant and provide the Participant Information Sheet digitally via QR code. Obtain informed consent through Nettskjema.
            \item Provide a second QR code that links to the Participant Demographic Information form on Nettskjema, which participants will complete before interacting with the robot.
        \end{itemize} 
    \vspace{0.2cm}
    \item \textbf{Robot Interaction:} The participant will interact with the robot, providing information about the university and local events. \newline
    \item \textbf{Post-Interaction Assessment:} The participant will complete the feedback form, focusing on selected aspects of the SecuRoPS framework. \newline
    \item \textbf{Feedback Collection:} Collect and anonymize the data. \newline
    \item \textbf{Data Handling:} All data will be securely stored in Nettskjema and Østfold University 
    Office 365 OneDrive, and will be used exclusively for research purposes.
\end{enumerate}
\subsection*{Ethical Considerations}
\begin{itemize}
    \item Maintain participant confidentiality.
    \item Ensure transparency in the robot’s functionalities and data collection practices.
\end{itemize}

\section{Post-Interaction Assessment Survey}\label{secD}
\begin{enumerate}
    \item Based on your experience and interaction with the robot ARI, do you perceive physical safety as one of the main characteristics of the robot? $\Box$	Yes, 		$\Box$  No 	\\ 
   % \vspace{0.3cm}
    If so, where?  \rule{11.2cm}{0.5pt} \\
 %   \vspace{0.3cm}
    If not, how can the physical safety of users be improved? \rule{4cm}{0.5pt} \\
    \rule{13.4cm}{0.5pt} 
    \item Based on your interaction and experience with the robot and the information provided, do you feel that your privacy is at risk due to ARI’s data handling? \\ $\Box$	Yes, 		$\Box$  No 	\\
%     \vspace{0.3cm}
    If so, why?  \rule{11.5cm}{0.5pt} \\
   %  \vspace{0.3cm}
    If not, how do you think the robot’s data handling could be better managed? 
    \rule{13.4cm}{0.5pt} 
    \item Assume that ARI’s design prevents access through WiFi or any other wireless means. Do you think the robot's implementors adopted ways to prevent it from physical tampering? Take a look at ARI, including from behind. $\Box$	Yes, 		$\Box$  No 	\\
     If so, how?  \rule{11.5cm}{0.5pt} \\
   %  \vspace{0.3cm}
    If not, how can the physical non-tampering of the robot be enhanced?   \rule{1.7cm}{0.5pt} \\
    \rule{13.4cm}{0.5pt} 

    \item Based on your interaction with ARI, can you see any perceived cybersecurity vulnerabilities in the robot design that could be exploited during interaction with users? $\Box$	Yes, 		$\Box$  No 	\\
     If so, how?  \rule{11.5cm}{0.5pt} \\
      If not, which cybersecurity vulnerabilities should be addressed?   \rule{2.6cm}{0.5pt} \\
    \rule{13.4cm}{0.5pt} 
    \item Based on your experience and interaction with ARI, do you think the robot has an acceptable interface usability?  $\Box$	Yes, 		$\Box$  No 	\\
     If so, how?  \rule{11.5cm}{0.5pt} \\
      If not, how can the usability of this robot be improved?   \rule{3.85cm}{0.5pt} \\
    \rule{13.4cm}{0.5pt} 
    
    \item Based on your interaction with the robot, do you think it accommodates diverse user needs in terms of accessibility?  $\Box$	Yes, 		$\Box$  No 	\\
    If so, how?  \rule{11.5cm}{0.5pt} \\
    If not, how can the usability of this robot be improved?   \rule{3.85cm}{0.5pt} \\
    \rule{13.4cm}{0.5pt} 

    \item  Based on your experience with the robot today, do you think ARI demonstrates ethical behaviour (e.g. giving misleading or biased information) and transparency in data collection? $\Box$	Yes, 		$\Box$  No 	\\
    If so, how?  \rule{11.5cm}{0.5pt} \\
    If not, how can compliance with ethical guidelines and transparency in data collection be improved?   \rule{9.3cm}{0.5pt} \\
    \rule{13.4cm}{0.5pt} 
    
    \item Based on your interaction with the robot, do you think the robot is complying with applicable privacy laws like GDPR?  $\Box$	Yes, 		$\Box$  No 	\\
     If so, how?  \rule{11.5cm}{0.5pt} \\
    If not, how can legal compliance be improved?   \rule{5.45cm}{0.5pt} \\
    \rule{13.4cm}{0.5pt} 
\end{enumerate}

\end{appendices}
\newpage
\bibliography{sn-bibliography}% common bib file

\end{document}